\documentclass[runningheads]{llncs}
\usepackage[T1]{fontenc}
\usepackage{subcaption}
\usepackage{tabularx}
\usepackage{makecell}
\usepackage{float}
\usepackage{amsmath}

\usepackage{booktabs}
\usepackage{multirow}
\usepackage{siunitx}
\usepackage{xcolor} 
\usepackage{amsfonts}
\usepackage{url}

\usepackage{graphicx,verbatim}
\usepackage{color}
\usepackage{hyperref}
\hypersetup{
    colorlinks=true,
    linkcolor=blue,
    citecolor=blue,
    urlcolor=blue
}

\begin{document}
\title{Joint Imaging–ROI Representation Learning via Cross-View Contrastive Alignment for Brain Disorder Classification}
%

\author{Wei Liang \and Lifang He}
\authorrunning{W. Liang and L. He}

\institute{Lehigh University, Bethlehem, PA, USA\\
\email{\{wel425, lih319\}@lehigh.edu}}
\maketitle              
\begin{abstract}
Brain imaging classification is commonly approached from two perspectives: modeling the full image volume to capture global anatomical context, or constructing ROI-based graphs to encode localized and topological interactions. Although both representations have demonstrated independent efficacy, their relative contributions and potential complementarity remain insufficiently understood. Existing fusion approaches are typically task-specific and do not enable controlled evaluation of each representation under consistent training settings. To address this gap, we propose a unified cross-view contrastive framework for joint imaging–ROI representation learning. Our method learns subject-level global (imaging) and local (ROI-graph) embeddings and aligns them in a shared latent space using a bidirectional contrastive objective, encouraging representations from the same subject to converge while separating those from different subjects. This alignment produces comparable embeddings suitable for downstream fusion and enables systematic evaluation of imaging-only, ROI-only, and joint configurations within a unified training protocol. Extensive experiments on the ADHD-200 and ABIDE datasets demonstrate that joint learning consistently improves classification performance over either branch alone across multiple backbone choices. Moreover, interpretability analyses reveal that imaging-based and ROI-based branches emphasize distinct yet complementary discriminative patterns, explaining the observed performance gains. These findings provide principled evidence that explicitly integrating global volumetric and ROI-level representations is a promising direction for neuroimaging-based brain disorder classification. The source code is available at \url{https://anonymous.4open.science/r/imaging-roi-contrastive-152C/}.


\end{abstract}
\section{Introduction}
\label{sec:intro}
Neuroimaging has become an essential tool for studying neurological and neurodevelopmental disorders, enabling data-driven diagnosis and characterization of disease-related brain alterations~\cite{dong2025exploring,zhang2020survey}. With advances in deep learning, brain imaging classification methods have increasingly focused on learning discriminative representations directly from volumetric data~\cite{abrol2021deep}. Despite substantial progress, a fundamental question remains open: \textbf{how should brain imaging data be represented for optimal classification performance?}

Existing approaches largely follow two distinct paradigms. The first models the entire 3D imaging volume using convolutional or transformer-based architectures to capture global anatomical context~\cite{khvostikov20183dcnnbasedclassificationusing,kumar2025ocnn,qiang2025classification}. These methods leverage holistic spatial information but may overlook fine-grained inter-regional relationships.
 The second paradigm constructs region-of-interest (ROI) graphs, where nodes correspond to predefined brain regions and edges encode structural or functional relationships~\cite{cui2022braingb,luo2024graph,pini2025connectivity,said2023neurograph}. ROI-based graph modeling emphasizes localized topology and inter-regional interactions that are often clinically meaningful. While both paradigms have demonstrated effectiveness in various brain disorder classification tasks~\cite{alp2024joint,dong2025exploring}, they capture fundamentally different aspects of brain organization.

Although imaging-based and ROI-based representations have been studied extensively in isolation~\cite{JOMEIRI2025111065,luo2024graph,qiang2025classification}, their relative contributions and potential complementarity remain poorly characterized. Prior fusion approaches typically integrate modalities within customized architectures tailored to specific tasks~\cite{dong2025multi,mao2024brain}. Such designs often confound architectural differences with representational benefits, making it difficult to determine whether performance improvements stem from genuine representational synergy or from variations in model complexity and training strategies. A principled and controlled evaluation framework is therefore needed to clarify how global and ROI-level representations contribute individually and jointly to classification performance.


To address this limitation, we introduce a unified framework for joint imaging–ROI representation learning via cross-view contrastive alignment. Our approach employs two modular encoders to extract subject-level global imaging embeddings and local ROI-graph embeddings. A bidirectional cross-view contrastive objective aligns these heterogeneous representations in a shared latent space, encouraging consistency across views while preserving discriminative capacity within each branch. This alignment yields comparable embeddings suitable for downstream fusion, enabling systematic comparison of imaging-only, ROI-only, and joint configurations under identical training conditions. 

Extensive experiments on the ADHD-200 and ABIDE datasets demonstrate that joint learning consistently outperforms single-branch baselines across multiple encoder choices. Beyond quantitative gains, interpretability analyses reveal that the two branches focus on distinct yet complementary neuroanatomical patterns, providing insight into their synergistic effects. Collectively, our findings establish that explicitly aligning and integrating global volumetric and ROI-level graph representations offers a principled and effective strategy for brain disorder classification. 
Our contributions are threefold:
\begin{itemize}
    \item We propose a unified cross-view contrastive framework for joint modeling of volumetric imaging and ROI-based graph representations under consistent training settings.
    \item We provide a systematic and controlled evaluation of imaging-only, ROI-only, and joint configurations, clarifying their individual and complementary contributions.
    \item Through experiments and interpretability analyses, we demonstrate that aligned imaging–ROI fusion yields consistent and complementary benefits for brain disorder classification.
\end{itemize}

\section{Method}
\label{sec:method}
\begin{figure*}[t]
  \centering
  \includegraphics[width=\textwidth]{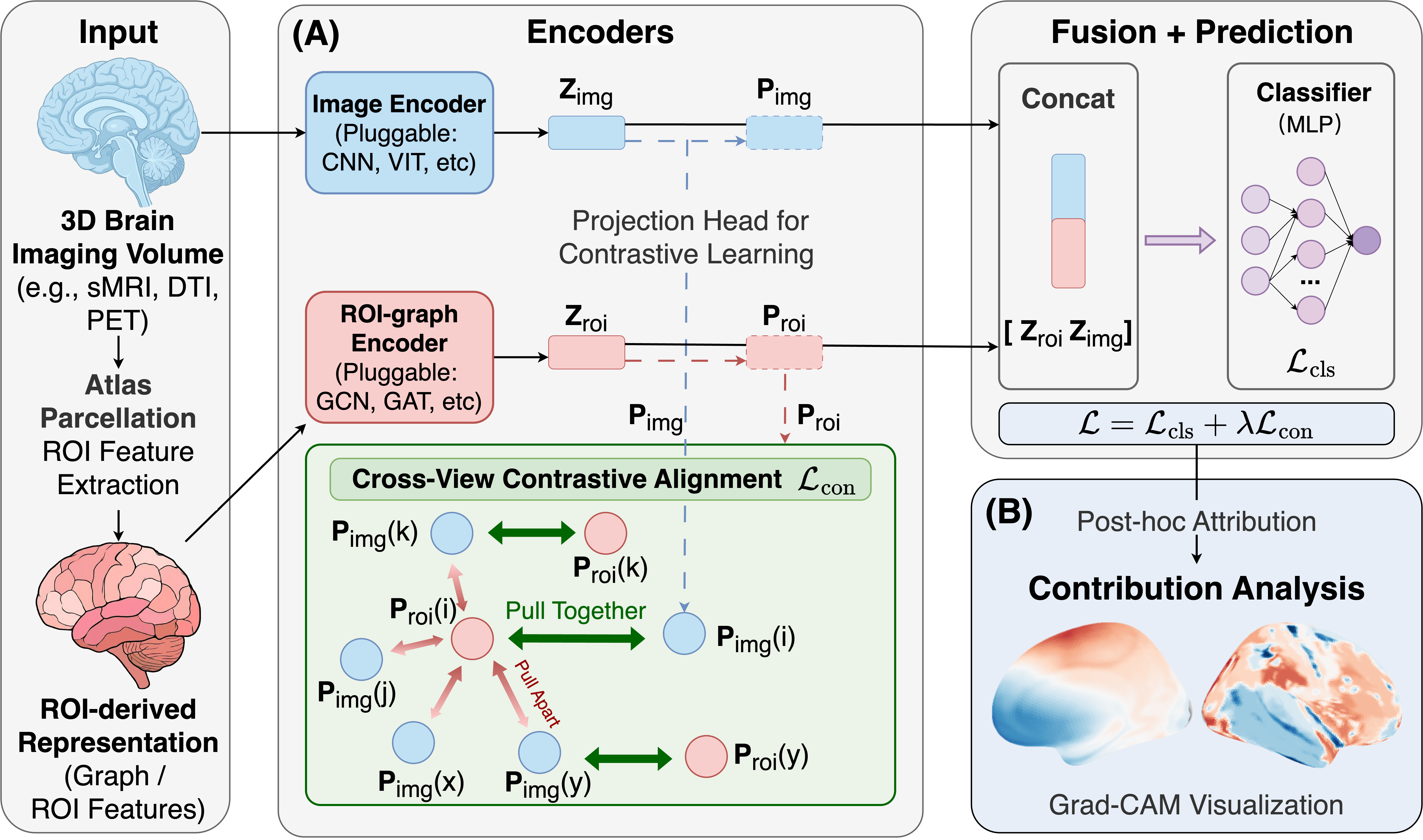}  
  \caption{\textbf{Overview of the proposed joint imaging--ROI representation learning framework.} 
  (A) Joint representation learning via cross-view contrastive alignment. Subject-level global imaging and local ROI-graph embeddings are extracted by modular encoders and aligned in a shared latent space to enable downstream fusion and classification.
(B) Post-hoc contribution analysis. Branch-specific attribution maps quantify complementary discriminative patterns captured by the imaging and ROI-graph representations.
}
  \label{fig:method_overview}
\end{figure*}

We propose a unified framework for joint imaging–ROI representation learning through cross-view contrastive alignment. As illustrated in Fig.~\ref{fig:method_overview}, the framework consists of three components: (i) extraction of global imaging and local ROI-graph representations, (ii) cross-view contrastive alignment in a shared latent space, and (iii) downstream fusion and classification.


\subsection{Imaging and ROI-Graph Representation Learning}
Given a subject-level imaging volume $\mathbf{x}_i$, we construct two complementary representations: a global volumetric representation and a local ROI-graph representation.
The imaging embedding is obtained via an encoder $f_{\mathrm{img}}(\cdot)$:
\begin{equation}
\mathbf{z}^{(i)}_{\mathrm{img}} = f_{\mathrm{img}}(\mathbf{x}_i), \qquad \text{ where }\mathbf{z}^{(i)}_{\mathrm{img}} \in \mathbb{R}^{d_{\mathrm{img}}}.
\end{equation}

To model inter-regional relationships, we construct an ROI graph $\mathcal{G}(\mathbf{x}_i)$ using AAL atlas parcellation~\cite{tzourio2002automated}. Node features correspond to mean voxel intensities within each ROI, and edges are defined by pairwise Pearson correlations between regions, yielding a subject-specific adjacency matrix. The ROI embedding is then computed via a graph encoder $f_{\mathrm{roi}}(\cdot)$:
\begin{equation}
\mathbf{z}^{(i)}_{\mathrm{roi}} = f_{\mathrm{roi}}(\mathcal{G}(\mathbf{x}_i)), \qquad \text{ where } \mathbf{z}^{(i)}_{\mathrm{roi}} \in \mathbb{R}^{d_{\mathrm{roi}}}.
\end{equation}

In practice, we instantiate $f_{\mathrm{img}}$ using 3DSC-TF~\cite{qiang2025classification}, a hybrid CNN–Transformer architecture designed for 3D brain imaging, and $f_{\mathrm{roi}}$ using NeuroGraph~\cite{said2023neurograph}, a graph neural network framework developed for brain connectome analysis. These models serve as strong and representative backbones for volumetric and ROI-based learning, respectively. Importantly, our framework is modular and agnostic to specific encoder choices. Alternative backbone choices are evaluated in Sec.~\ref{sec:experiments}.

\subsection{Cross-View Contrastive Alignment}
To align heterogeneous representations from imaging and ROI views, we adopt a bidirectional InfoNCE objective for cross-view contrastive learning~\cite{chen2020simple}. Given subject-level imaging and ROI embeddings $\mathbf{z}^{(i)}_{\text{img}}$ and $\mathbf{z}^{(i)}_{\text{roi}}$, two projection heads $g_{\text{img}}(\cdot)$ and $g_{\text{roi}}(\cdot)$ map them into a shared latent space:
\begin{equation}
\mathbf{p}^{(i)}_{\text{img}} = g_{\text{img}}(\mathbf{z}^{(i)}_{\text{img}}), \qquad
\mathbf{p}^{(i)}_{\text{roi}} = g_{\text{roi}}(\mathbf{z}^{(i)}_{\text{roi}}).
\end{equation}
For a mini-batch of size $B$, we compute cross-view similarities as:
\begin{equation}
S_{ij}=\frac{\text{sim}\!\left(\mathbf{p}^{(i)}_{\mathrm{img}},\,\mathbf{p}^{(j)}_{\mathrm{roi}}\right)}{\tau},
\end{equation}
where $\tau$ is a temperature parameter and $\text{sim}(\cdot, \cdot)$ denotes cosine similarity.

Using these similarities, the bidirectional InfoNCE loss~\cite{chen2020simple} is defined as:
\begin{equation}
\mathcal{L}_{\mathrm{con}}
= -\frac{1}{2B}\sum_{i=1}^{B}\left[
\log\frac{\exp(S_{ii})}{\sum_{j=1}^{B}\exp(S_{ij})}
+
\log\frac{\exp(S_{ii})}{\sum_{j=1}^{B}\exp(S_{ji})}
\right].
\end{equation}
where $S_{ii}$ corresponds to positive cross-view pairs from the same subject, and $S_{ij}$ for $i\neq j$ correspond to negative pairs across different subjects.
This formulation promotes cross-view alignment of representations from the same subject while preserving inter-subject discrimination.

\subsection{Fusion and Classification Objective}
The aligned embeddings are concatenated to form a joint representation:
$$\mathbf{z}^{(i)}_{\mathrm{fuse}}=\bigl[\mathbf{z}^{(i)}_{\mathrm{img}};\mathbf{z}^{(i)}_{\mathrm{roi}}\bigr].$$
A classifier $h(\cdot)$ maps the fused representation to class logits $\hat{y}_i$. The overall training objective combines the cross-entropy classification loss and the contrastive loss:
\begin{equation}
\mathcal{L} = \mathcal{L}_{\mathrm{cls}} + \lambda\,\mathcal{L}_{\mathrm{con}},
\end{equation}
where $\mathcal{L}_{\mathrm{cls}}$ is the standard cross-entropy loss for classification, and $\lambda$ controls the trade-off between discriminative supervision and cross-view alignment.


\section{Experiments and Results}
\label{sec:experiments}
We evaluate the proposed joint imaging–ROI framework along four dimensions: (i) comparison between single-branch (imaging-only, ROI-only) and joint learning; (ii) ablations on encoder choices and fusion mechanisms; (iii) robustness under missing-view settings; and (iv) interpretability analyses to quantify branch-specific contributions and complementarity.

\noindent\textbf{Dataset Description} We conducted experiments on two public structural MRI (sMRI) datasets: ADHD-200~\cite{10.3389/fnsys.2012.00062} and the Autism Brain Imaging Data Exchange (ABIDE)~\cite{tyszka2014largely}. ADHD-200 includes 776 subjects (ADHD vs. typically developing controls, TDC), and ABIDE contains 1112 subjects (autism spectrum disorder vs. TDC). Performance is evaluated using accuracy (Acc), area under the ROC curve (AUC), and F1-score (F1). All results are reported as mean~$\pm$~standard deviation over 5-fold stratified cross-validation.


\noindent\textbf{Single-Branch vs. Joint Learning} To establish strong single-branch baselines and evaluate the unified framework, we employ three 3D imaging encoders (ViT3D*~\cite{kunanbayev2024training}, RAE-ViT~\cite{JOMEIRI2025111065}, and 3DSC-TF~\cite{qiang2025classification}), together with two ROI-based models (DNN~\cite{chen2019multichannel} and NeuroGraph~\cite{said2023neurograph}). Comparisons are conducted across ROI-only, imaging-only, and joint configurations. 
As shown in Table~\ref{tab:combined_results}, joint learning consistently outperforms both imaging-only and ROI-only baselines on ADHD-200 and ABIDE across backbone choices. These improvements indicate that global volumetric features and ROI-level graph representations capture complementary information. Importantly, the performance gains persist across multiple imaging encoders, suggesting that the proposed framework is backbone-agnostic rather than tailored to a specific architecture. These observations motivate the encoder selection for subsequent fusion ablations.

\begin{table}[h]
\centering
\caption{Single-branch vs.\ joint learning performance on ADHD-200 and ABIDE.} 
\label{tab:combined_results}

\scriptsize
\setlength{\tabcolsep}{5pt}
\renewcommand{\arraystretch}{1.15}

\sisetup{
  table-number-alignment = center,
  detect-weight = true,
  separate-uncertainty = true,
  uncertainty-separator = {\,\pm\,}
}

\resizebox{\linewidth}{!}{
\begin{tabular}{@{} l S[table-format=2.2(2.2)] S[table-format=2.2(2.2)] S[table-format=2.2(2.2)] @{}}
\toprule
\textbf{Model} & {\textbf{Acc}} & {\textbf{AUC}} & {\textbf{F1}} \\
\midrule

& \multicolumn{3}{c}{\textbf{ADHD}} \\
\cmidrule(lr){2-4}

\multicolumn{4}{@{}l}{\textbf{\textit{ROI-only}}} \\
DNN        & \num{64.77 +- 4.88} & \num{67.73 +- 5.55} & \num{63.40 +- 5.60} \\
NeuroGraph & \num{63.48 +- 2.74} & \num{65.51 +- 3.23} & \num{50.25 +- 6.62} \\
\addlinespace[2pt]

\multicolumn{4}{@{}l}{\textbf{\textit{Imaging-only}}} \\
ViT3D*   & \num{66.84 +- 2.63} & \num{66.57 +- 6.19} & \num{64.62 +- 4.21} \\
RAE-ViT  & \num{64.26 +- 4.05} & \num{63.99 +- 7.37} & \num{61.75 +- 5.25} \\
3DSC-TF  & \num{68.65 +- 4.99} & \num{70.92 +- 4.93} & \num{68.42 +- 4.85} \\
\addlinespace[2pt]

\multicolumn{4}{@{}l}{\textbf{\textit{Joint}}} \\
NeuroGraph + ViT3D*    & \num{68.13 +- 1.98} & \num{69.70 +- 2.49} & \num{67.25 +- 1.72} \\
NeuroGraph + RAE-ViT   & \num{67.35 +- 4.86} & \num{69.76 +- 7.72} & \num{66.85 +- 5.24} \\
\textbf{NeuroGraph + 3DSC-TF} & \bfseries \num{69.29 +- 4.44} & \bfseries \num{72.73 +- 4.17} & \bfseries \num{69.01 +- 4.51} \\

\midrule

& \multicolumn{3}{c}{\textbf{ABIDE}} \\
\cmidrule(lr){2-4}

\multicolumn{4}{@{}l}{\textbf{\textit{ROI-only}}} \\
DNN        & \num{57.24 +- 4.44} & \num{62.84 +- 3.90} & \num{56.02 +- 4.93} \\
NeuroGraph & \num{61.09 +- 1.23} & \num{60.90 +- 3.41} & \num{59.52 +- 1.26} \\
\addlinespace[2pt]

\multicolumn{4}{@{}l}{\textbf{\textit{Imaging-only}}} \\
ViT3D*   & \num{56.94 +- 0.95} & \num{54.51 +- 2.38} & \num{55.57 +- 1.95} \\
RAE-ViT  & \num{57.95 +- 2.51} & \num{56.78 +- 2.06} & \num{55.75 +- 3.43} \\
3DSC-TF  & \num{59.17 +- 2.39} & \num{57.91 +- 2.16} & \num{58.33 +- 3.12} \\
\addlinespace[2pt]

\multicolumn{4}{@{}l}{\textbf{\textit{Joint}}} \\
NeuroGraph + ViT3D*   & \num{58.25 +- 2.68} & \num{58.15 +- 3.49} & \num{57.04 +- 3.62} \\
NeuroGraph + RAE-ViT  & \num{60.38 +- 3.12} & \num{60.44 +- 4.56} & \num{59.55 +- 2.75} \\
\textbf{NeuroGraph + 3DSC-TF} & \bfseries \num{62.54 +- 1.79} & \bfseries \num{64.08 +- 2.14} & \bfseries \num{61.71 +- 1.43} \\

\bottomrule
\end{tabular}}
\vspace{-15pt}
\end{table}
\noindent\textbf{Encoder and Fusion Ablation} To further analyze architectural choices, we fix the imaging encoder to the strongest-performing backbone (3DSC-TF) and evaluate two ROI encoders (DNN and NeuroGraph) under three fusion strategies: simple concatenation (Concat), bidirectional cross-attention (Cross-attn), and the proposed contrastive alignment (Contra).
Table~\ref{tab:fusion_ablation_combined} reveals two consistent findings. First, NeuroGraph consistently outperforms DNN across datasets and fusion mechanisms, indicating that graph-based message passing more effectively captures inter-regional connectivity than independent node-level features. Second, contrastive alignment achieves the best overall performance in most settings, demonstrating that explicitly aligning heterogeneous representations in a shared latent space produces more compatible embeddings for downstream classification.


\begin{table}[t]
\centering
\caption{Encoder and joint learning ablation on ADHD-200 and ABIDE.}
\label{tab:fusion_ablation_combined}

\footnotesize
\setlength{\tabcolsep}{3pt}
\renewcommand{\arraystretch}{1.15}

\sisetup{
  table-number-alignment = center,
  detect-weight = true,
  separate-uncertainty = true,
  uncertainty-separator = {\,\pm\,}
}

\resizebox{\linewidth}{!}{
\begin{tabular}{@{} l l l S[table-format=2.2(2.2)] S[table-format=2.2(2.2)] S[table-format=2.2(2.2)] @{}}
\toprule
\textbf{Imaging Encoder} & \textbf{ROI Encoder} & \textbf{Fusion} & {\textbf{Acc}} & {\textbf{AUC}} & {\textbf{F1}} \\
\midrule

& & & \multicolumn{3}{c}{\textbf{ADHD}} \\
\cmidrule(lr){4-6}

\multirow{6}{*}{3DSC-TF}
& \multirow{3}{*}{DNN} & Concat     & \num{59.87 +- 4.07} & \num{66.73 +- 3.56} & \num{58.62 +- 3.94} \\
&                      & Cross-attn & \num{60.65 +- 7.63} & \num{66.49 +- 4.62} & \num{59.11 +- 8.77} \\
&                      & Contra    & \num{62.84 +- 4.20} & \num{68.69 +- 4.62} & \num{61.06 +- 5.79} \\
\cmidrule(l){2-6}
& \multirow{3}{*}{NeuroGraph} & Concat     & \num{66.32 +- 4.53} & \num{65.94 +- 7.88} & \num{64.17 +- 6.23} \\
&                             & Cross-attn & \num{68.00 +- 4.23} & \num{69.53 +- 5.94} & \num{67.71 +- 4.32} \\
&                             & \textbf{Contra} & \bfseries \num{69.29 +- 4.44} & \bfseries \num{72.73 +- 4.17} & \bfseries \num{69.01 +- 4.51} \\

\midrule

& & & \multicolumn{3}{c}{\textbf{ABIDE}} \\
\cmidrule(lr){4-6}
\multirow{6}{*}{3DSC-TF} 
& \multirow{3}{*}{DNN} & Concat & {\num{56.24 +- 2.78}} & {\num{60.24 +- 3.15}} & {\num{51.61 +- 9.64}} \\
& & Cross-attn & \multicolumn{1}{c}{57.84 $\pm$ 6.60} & \multicolumn{1}{c}{59.35 $\pm$ 5.98} & \multicolumn{1}{c}{54.37 $\pm$ 4.09} \\
&                             & Contra    & \num{57.85 +- 4.16} & \num{62.71 +- 5.26} & \num{57.35 +- 4.04} \\
\cmidrule(l){2-6}
& \multirow{3}{*}{NeuroGraph} & Concat     & \num{60.89 +- 1.91} & \num{61.73 +- 4.47} & \num{57.85 +- 4.54} \\
&                             & Cross-attn & \num{61.70 +- 5.12} & \bfseries \num{64.63 +- 4.17} & \num{61.01 +- 4.83} \\
&                             & \textbf{Contra} & \bfseries \num{62.54 +- 1.79} & \num{64.08 +- 2.14} & \bfseries \num{61.71 +- 1.43} \\

\bottomrule
\end{tabular}}
\end{table}

\begin{figure}[t]
  \centering
  \includegraphics[width=\linewidth]{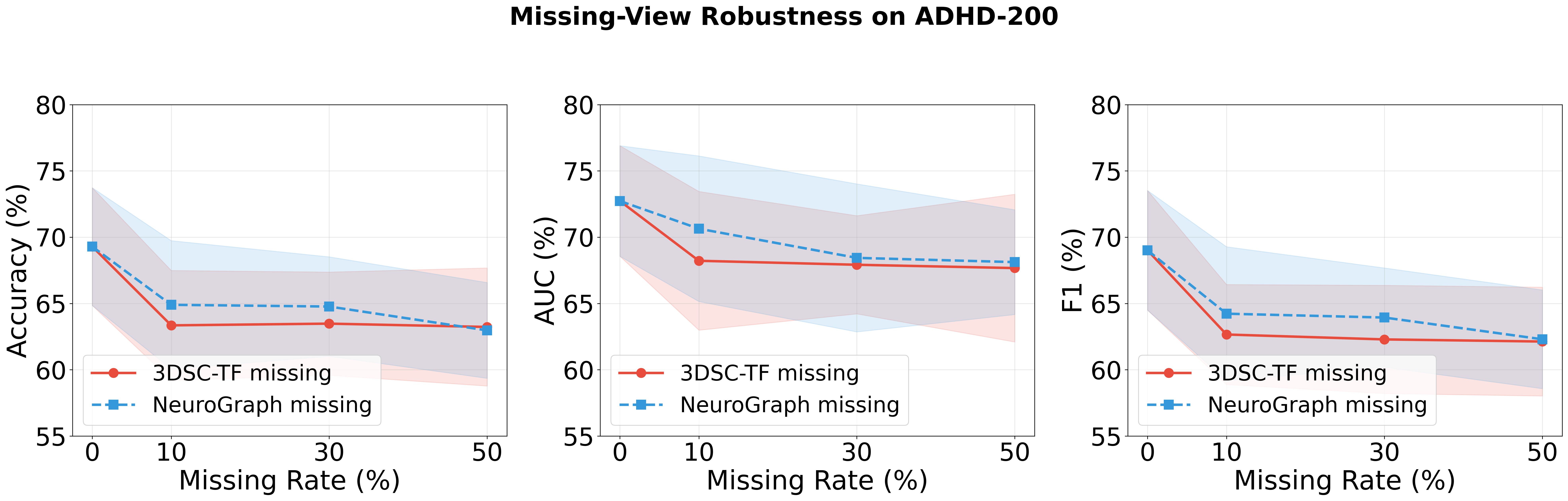}
  \caption{Missing-view robustness on ADHD-200. Shaded regions denote $\pm1$ std over five-fold cross-validation.}
  \label{fig:missing}
\end{figure}

\noindent\textbf{Missing-View Robustness.}
In practical clinical scenarios, imaging modalities or ROI-derived features may be unavailable due to acquisition or preprocessing failures \cite{gao2023multimodal,uchida2018smaller}. To assess robustness under missing views, we simulate view absence by zero-masking one branch for a random subset of subjects at rates of 10\%, 30\%, and 50\% on the ADHD-200 cohort, applied consistently during both training and testing. As illustrated in Fig.~\ref{fig:missing}, masking either branch leads to an initial performance drop compared to the full-view setting. However, increasing the missing rate from 10\% to 50\% results in only moderate additional degradation. The bounded decline suggests that per-branch supervision preserves individual discriminative capacity, while cross-view contrastive alignment facilitates implicit knowledge transfer between branches. Consequently, the remaining view can partially compensate for the missing modality.

\begin{figure}[t]
    \centering
    \includegraphics[width=\linewidth]{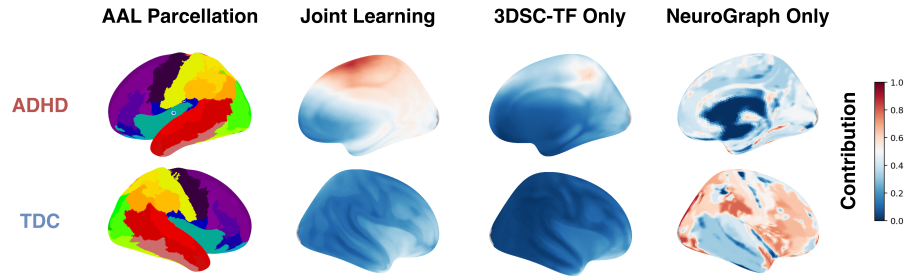}
    \caption{Contribution maps for ADHD and TDC. The AAL parcellation (first column) provides anatomical reference; remaining columns show normalized contribution scores (0--1) for Joint Learning, 3DSC-TF Only, and NeuroGraph Only.}
    \label{fig:contribution}
\end{figure}

\noindent\textbf{Contribution Analysis} To interpret learned decision patterns, we apply Grad-CAM~\cite{fayyaz2025grad} to derive region-level contribution maps from the test sets across 5-fold cross-validation on the ADHD-200 cohort, as shown in Fig.~\ref{fig:contribution}. Such attribution visualizations are widely adopted in medical imaging to assess clinical face validity~\cite{zhou2022interpretable}. Compared with unimodal models, the joint framework produces more spatially coherent contribution maps. The imaging-only branch yields relatively diffuse patterns, whereas the ROI-only branch produces sharper but more heterogeneous activations. In contrast, the joint model highlights regions consistently supported across both representations, suggesting synergistic integration.

Regions with higher contributions in the joint model predominantly involve frontal, sensorimotor, orbitofrontal, and limbic systems, which have been repeatedly implicated in ADHD-200 by large-scale and meta-analytic neuroimaging studies. For example, prior meta-analyses report ADHD-200-related alterations in the superior frontal, precentral, and orbitofrontal cortices~\cite{chen2025multimodal}, as well as volumetric differences in limbic structures such as the hippocampus~\cite{hoogman2017subcortical}. Functional evidence further implicates orbitofrontal and supplementary motor regions in reward processing and inhibitory control~\cite{hart2013meta,kringelbach2005human,nachev2008functional}. While attribution maps do not establish causal neurobiological mechanisms, the alignment between model-sensitive regions and established ADHD-related circuits supports the clinical plausibility of the learned representations and clarifies how the two branches contribute jointly.

\section{Conclusion}
\label{sec:conclusion}
In this work, we proposed a unified cross-view contrastive framework for joint imaging–ROI representation learning in brain disorder classification. By aligning heterogeneous embeddings in a shared latent space, the framework enables principled integration of volumetric and ROI-level graph representations. Experiments on ADHD-200 and ABIDE demonstrate consistent improvements over single-branch baselines across multiple backbones. Interpretability analyses further reveal complementary neuroanatomical patterns captured by the two representations. These results suggest that cross-view alignment provides an effective strategy for integrating complementary brain representations in neuroimaging-based diagnosis.


\noindent\textbf{Acknowledgment} This work was supported in part by NIH (R01LM013519, RF1AG077820), NSF (IIS-2319451, MRI-2215789), DOE (DE-SC0025801), and Lehigh University (CORE and RIG). 

\noindent\textbf{Disclosure of Interests} The authors have no competing interests to declare
that are relevant to the content of this article.

\bibliographystyle{splncs04}
\bibliography{refs}

@article{dong2025exploring,
  title={Exploring the impact of APOE $\varepsilon$4 on functional connectivity in Alzheimer’s disease across cognitive impairment levels},
  author={Dong, Kangli and Liang, Wei and Hou, Ting and Lu, Zhijie and Hao, Yixuan and Li, Chenrui and Qiu, Yue and Kong, Nan and Cheng, Yan and Wen, Yaqi and others},
  journal={NeuroImage},
  volume={305},
  pages={120951},
  year={2025},
  publisher={Elsevier}
}

@article{zhang2020survey,
  title={A survey on deep learning for neuroimaging-based brain disorder analysis},
  author={Zhang, Li and Wang, Mingliang and Liu, Mingxia and Zhang, Daoqiang},
  journal={Frontiers in neuroscience},
  volume={14},
  pages={779},
  year={2020},
  publisher={Frontiers Media SA}
}

@article{luo2024graph,
  title={Graph neural networks for brain graph learning: A survey},
  author={Luo, Xuexiong and Wu, Jia and Yang, Jian and Xue, Shan and Beheshti, Amin and Sheng, Quan Z and McAlpine, David and Sowman, Paul and Giral, Alexis and Yu, Philip S},
  journal={arXiv preprint arXiv:2406.02594},
  year={2024}
}

@misc{khvostikov20183dcnnbasedclassificationusing,
      title={3D CNN-based classification using sMRI and MD-DTI images for Alzheimer disease studies}, 
      author={Alexander Khvostikov and Karim Aderghal and Jenny Benois-Pineau and Andrey Krylov and Gwenaelle Catheline},
      year={2018},
      eprint={1801.05968},
      archivePrefix={arXiv},
      primaryClass={cs.CV},
}

@article{JOMEIRI2025111065,
title = {Regional attention-enhanced vision transformer for accurate Alzheimer's disease classification using sMRI data},
journal = {Computers in Biology and Medicine},
volume = {197},
pages = {111065},
year = {2025},
issn = {0010-4825},
author = {Alireza Jomeiri and Ahmad {Habibizad Navin} and Mahboubeh Shamsi},
}

@ARTICLE{10.3389/fnsys.2012.00062,
    
AUTHOR={Milham, Michael P. and Fair, Damien  and Mennes, Maarten  and Mostofsky, Stewart H.},
           
TITLE={The adhd-200 consortium: a model to advance the translational potential of neuroimaging in clinical neuroscience},
          
JOURNAL={Frontiers in Systems Neuroscience},
          
VOLUME={Volume 6 - 2012},
  
YEAR={2012},
  
ISSN={1662-5137},
  
}

@inproceedings{kumar2025ocnn,
  title={OCNN Classification Model Framework: A System for Classifying Alzheimers Disease from sMRI Images},
  author={Kumar, V Santhosh and Lanitha, B},
  booktitle={2025 8th International Conference on Computing Methodologies and Communication (ICCMC)},
  pages={620--627},
  year={2025},
  organization={IEEE}
}

@article{qiang2025classification,
  title={Classification of Alzheimer’s disease by jointing 3D depthwise separable convolutional neural network and transformer},
  author={Qiang, Yan-Rui and Zhou, Qin-Yi and Li, Jia-Ni and Xie, Ming-Yu and Cui, Xiaodong and Zhang, Shao-Wu},
  journal={Expert Systems with Applications},
  volume={286},
  pages={127720},
  year={2025},
  publisher={Elsevier}
}

@inproceedings{kunanbayev2024training,
  title={Training ViT with limited data for Alzheimer’s disease classification: an empirical study},
  author={Kunanbayev, Kassymzhomart and Shen, Vyacheslav and Kim, Dae-Shik},
  booktitle={International Conference on Medical Image Computing and Computer-Assisted Intervention},
  pages={334--343},
  year={2024},
  organization={Springer}
}

@article{pini2025connectivity,
  title   = {Can brain network connectivity facilitate the clinical development of disease-modifying anti-Alzheimer drugs?},
  author  = {Pini, Lorenzo and Lista, Simone and Griffa, Alessandra and Allali, Gilles and Imbimbo, Bruno P.},
  journal = {Brain Communications},
  volume  = {7},
  number  = {1},
  pages   = {fcae460},
  year    = {2025},
}

@article{chen2019multichannel,
  title={A multichannel deep neural network model analyzing multiscale functional brain connectome data for attention deficit hyperactivity disorder detection},
  author={Chen, Ming and Li, Hailong and Wang, Jinghua and Dillman, Jonathan R and Parikh, Nehal A and He, Lili},
  journal={Radiology: Artificial Intelligence},
  volume={2},
  number={1},
  pages={e190012},
  year={2019},
  publisher={Radiological Society of North America}
}

@article{said2023neurograph,
  title={Neurograph: Benchmarks for graph machine learning in brain connectomics},
  author={Said, Anwar and Bayrak, Roza and Derr, Tyler and Shabbir, Mudassir and Moyer, Daniel and Chang, Catie and Koutsoukos, Xenofon},
  journal={Advances in Neural Information Processing Systems},
  volume={36},
  pages={6509--6531},
  year={2023}
}

@article{tzourio2002automated,
  title={Automated anatomical labeling of activations in SPM using a macroscopic anatomical parcellation of the MNI MRI single-subject brain},
  author={Tzourio-Mazoyer, Nathalie and Landeau, Brigitte and Papathanassiou, Dimitri and Crivello, Fabrice and Etard, Octave and Delcroix, Nicolas and Mazoyer, Bernard and Joliot, Marc},
  journal={Neuroimage},
  volume={15},
  number={1},
  pages={273--289},
  year={2002},
  publisher={Elsevier}
}

@inproceedings{chen2020simple,
  title={A simple framework for contrastive learning of visual representations},
  author={Chen, Ting and Kornblith, Simon and Norouzi, Mohammad and Hinton, Geoffrey},
  booktitle={International conference on machine learning},
  pages={1597--1607},
  year={2020},
  organization={PmLR}
}

@article{abrol2021deep,
  title={Deep learning encodes robust discriminative neuroimaging representations to outperform standard machine learning},
  author={Abrol, Anees and Fu, Zening and Salman, Mustafa and Silva, Rogers and Du, Yuhui and Plis, Sergey and Calhoun, Vince},
  journal={Nature communications},
  volume={12},
  number={1},
  pages={353},
  year={2021},
  publisher={Nature Publishing Group UK London}
}

@article{cui2022braingb,
  title={Braingb: a benchmark for brain network analysis with graph neural networks},
  author={Cui, Hejie and Dai, Wei and Zhu, Yanqiao and Kan, Xuan and Gu, Antonio Aodong Chen and Lukemire, Joshua and Zhan, Liang and He, Lifang and Guo, Ying and Yang, Carl},
  journal={IEEE transactions on medical imaging},
  volume={42},
  number={2},
  pages={493--506},
  year={2022},
  publisher={IEEE}
}

@article{alp2024joint,
  title={Joint transformer architecture in brain 3D MRI classification: its application in Alzheimer’s disease classification},
  author={Alp, Sait and Akan, Taymaz and Bhuiyan, Md Shenuarin and Disbrow, Elizabeth A and Conrad, Steven A and Vanchiere, John A and Kevil, Christopher G and Bhuiyan, Mohammad AN},
  journal={Scientific Reports},
  volume={14},
  number={1},
  pages={8996},
  year={2024},
  publisher={Nature Publishing Group UK London}
}

@article{mao2024brain,
  title={Brain structural connectivity guided vision transformers for identification of functional connectivity characteristics in preterm neonates},
  author={Mao, Wei and Chen, Yuzhong and He, Zhibin and Wang, Zifan and Xiao, Zhenxiang and Sun, Yusong and He, Liang and Zhou, Jingchao and Guo, Weitong and Ma, Chong and others},
  journal={IEEE Journal of Biomedical and Health Informatics},
  volume={28},
  number={4},
  pages={2223--2234},
  year={2024},
  publisher={IEEE}
}

@article{dong2025multi,
  title={Multi-view brain network classification based on adaptive graph isomorphic information bottleneck Mamba},
  author={Dong, Changxu and Sun, Dengdi and Yu, Zhenda and Luo, Bin},
  journal={Expert Systems with Applications},
  volume={267},
  pages={126170},
  year={2025},
  publisher={Elsevier}
}

@article{tyszka2014largely,
  title={Largely typical patterns of resting-state functional connectivity in high-functioning adults with autism},
  author={Tyszka, J Michael and Kennedy, Daniel P and Paul, Lynn K and Adolphs, Ralph},
  journal={Cerebral cortex},
  volume={24},
  number={7},
  pages={1894--1905},
  year={2014},
  publisher={Oxford University Press}
}

@article{gao2023multimodal,
  title={Multimodal transformer network for incomplete image generation and diagnosis of Alzheimer’s disease},
  author={Gao, Xingyu and Shi, Feng and Shen, Dinggang and Liu, Manhua},
  journal={Computerized Medical Imaging and Graphics},
  volume={110},
  pages={102303},
  year={2023},
  publisher={Elsevier}
}

@article{uchida2018smaller,
  title={Smaller hippocampal volume and degraded peripheral hearing among Japanese community dwellers},
  author={Uchida, Yasue and Nishita, Yukiko and Kato, Takashi and Iwata, Kaori and Sugiura, Saiko and Suzuki, Hirokazu and Sone, Michihiko and Tange, Chikako and Otsuka, Rei and Ando, Fujiko and others},
  journal={Frontiers in Aging Neuroscience},
  volume={10},
  pages={319},
  year={2018},
  publisher={Frontiers Media SA}
}

@inproceedings{zhou2022interpretable,
  title={Interpretable graph convolutional network of multi-modality brain imaging for alzheimer’s disease diagnosis},
  author={Zhou, Houliang and He, Lifang and Zhang, Yu and Shen, Li and Chen, Brian},
  booktitle={2022 IEEE 19th International Symposium on Biomedical Imaging (ISBI)},
  pages={1--5},
  year={2022},
  organization={IEEE}
}

@article{fayyaz2025grad,
  title={Grad-CAM (Gradient-weighted Class Activation Mapping): A systematic literature review},
  author={Fayyaz, Abdul Muiz and Abdulkadir, Said Jadid and Talpur, Noureen and Al-Selwi, Safwan Mahmood and Hassan, Shahab Ul and Sumiea, Ebrahim Hamid},
  journal={Computers in Biology and Medicine},
  volume={198},
  pages={111200},
  year={2025},
  publisher={Elsevier}
}

@article{chen2025multimodal,
  title={A multimodal neuroimaging meta-analysis of functional and structural brain abnormalities in attention-deficit/hyperactivity disorder},
  author={Chen, Chao and Sun, Shilin and Chen, Ruoyi and Guo, Zixuan and Tang, Xinyue and Chen, Guanmao and Chen, Pan and Tang, Guixian and Huang, Li and Wang, Ying},
  journal={Progress in Neuro-Psychopharmacology and Biological Psychiatry},
  volume={136},
  pages={111199},
  year={2025},
  publisher={Elsevier}
}

@article{hoogman2017subcortical,
  title={Subcortical brain volume differences in participants with attention deficit hyperactivity disorder in children and adults: a cross-sectional mega-analysis},
  author={Hoogman, Martine and Bralten, Janita and Hibar, Derrek P and Mennes, Maarten and Zwiers, Marcel P and Schweren, Lizanne SJ and van Hulzen, Kimm JE and Medland, Sarah E and Shumskaya, Elena and Jahanshad, Neda and others},
  journal={The Lancet Psychiatry},
  volume={4},
  number={4},
  pages={310--319},
  year={2017},
  publisher={Elsevier}
}

@article{kringelbach2005human,
  title={The human orbitofrontal cortex: linking reward to hedonic experience},
  author={Kringelbach, Morten L},
  journal={Nature reviews neuroscience},
  volume={6},
  number={9},
  pages={691--702},
  year={2005},
  publisher={Nature Publishing Group UK London}
}

@article{nachev2008functional,
  title={Functional role of the supplementary and pre-supplementary motor areas},
  author={Nachev, Parashkev and Kennard, Christopher and Husain, Masud},
  journal={Nature reviews neuroscience},
  volume={9},
  number={11},
  pages={856--869},
  year={2008},
  publisher={Nature Publishing Group UK London}
}

@article{hart2013meta,
  title={Meta-analysis of functional magnetic resonance imaging studies of inhibition and attention in attention-deficit/hyperactivity disorder: exploring task-specific, stimulant medication, and age effects},
  author={Hart, Heledd and Radua, Joaquim and Nakao, Tomohiro and Mataix-Cols, David and Rubia, Katya},
  journal={JAMA psychiatry},
  volume={70},
  number={2},
  pages={185--198},
  year={2013},
  publisher={American Medical Association}
}
\end{document}